\def\year{2019}\relax
\documentclass[letterpaper]{article} %DO NOT CHANGE THIS
\usepackage{aaai19}  %Required
\usepackage{times}  %Required
\usepackage{helvet}  %Required
\usepackage{courier}  %Required
\usepackage{url}  %Required
\usepackage{graphicx}  %Required
\usepackage{booktabs}
\usepackage{amssymb}
\usepackage{amsmath}
\begin{document}
\title{Geometry-Aware Face Completion and Editing}
\author{Linsen Song, Jie Cao, Lingxiao Song, Yibo Hu, Ran He\thanks{Ran He is the corresponding author.}\\
National Laboratory of Pattern Recognition, CASIA \\
Center for Research on Intelligent Perception and Computing, CASIA \\
Center for Excellence in Brain Science and Intelligence Technology, CAS \\
University of Chinese Academy of Sciences, Beijing 100190, China\\
songlinsen2018@ia.ac.cn, \{jie.cao,yibo.hu\}@cripac.ia.ac.cn, \{lingxiao.song,rhe\}@nlpr.ia.ac.cn
}

\maketitle
\begin{abstract}
Face completion is a challenging generation task because it requires generating visually pleasing new pixels that are semantically consistent with the unmasked face region.  This paper proposes a geometry-aware Face Completion and Editing NETwork (FCENet) by systematically studying facial geometry from the unmasked region. Firstly, a facial geometry estimator is learned to estimate facial landmark heatmaps and parsing maps from the unmasked face image. Then, an encoder-decoder structure generator serves to complete a face image and disentangle its mask areas conditioned on both the masked face image and the estimated facial geometry images. Besides, since low-rank property exists in manually labeled masks, a low-rank regularization term is imposed on the disentangled masks, enforcing our completion network to manage occlusion area with various shape and size. Furthermore, our network can generate diverse results from the same masked input by modifying estimated facial geometry, which provides a flexible mean to edit the completed face appearance. Extensive experimental results qualitatively and quantitatively demonstrate that our network is able to generate visually pleasing face completion results and edit face attributes as well. 
\end{abstract}

\begin{figure}[htb]
  \centering
  \includegraphics[width=.8\columnwidth]{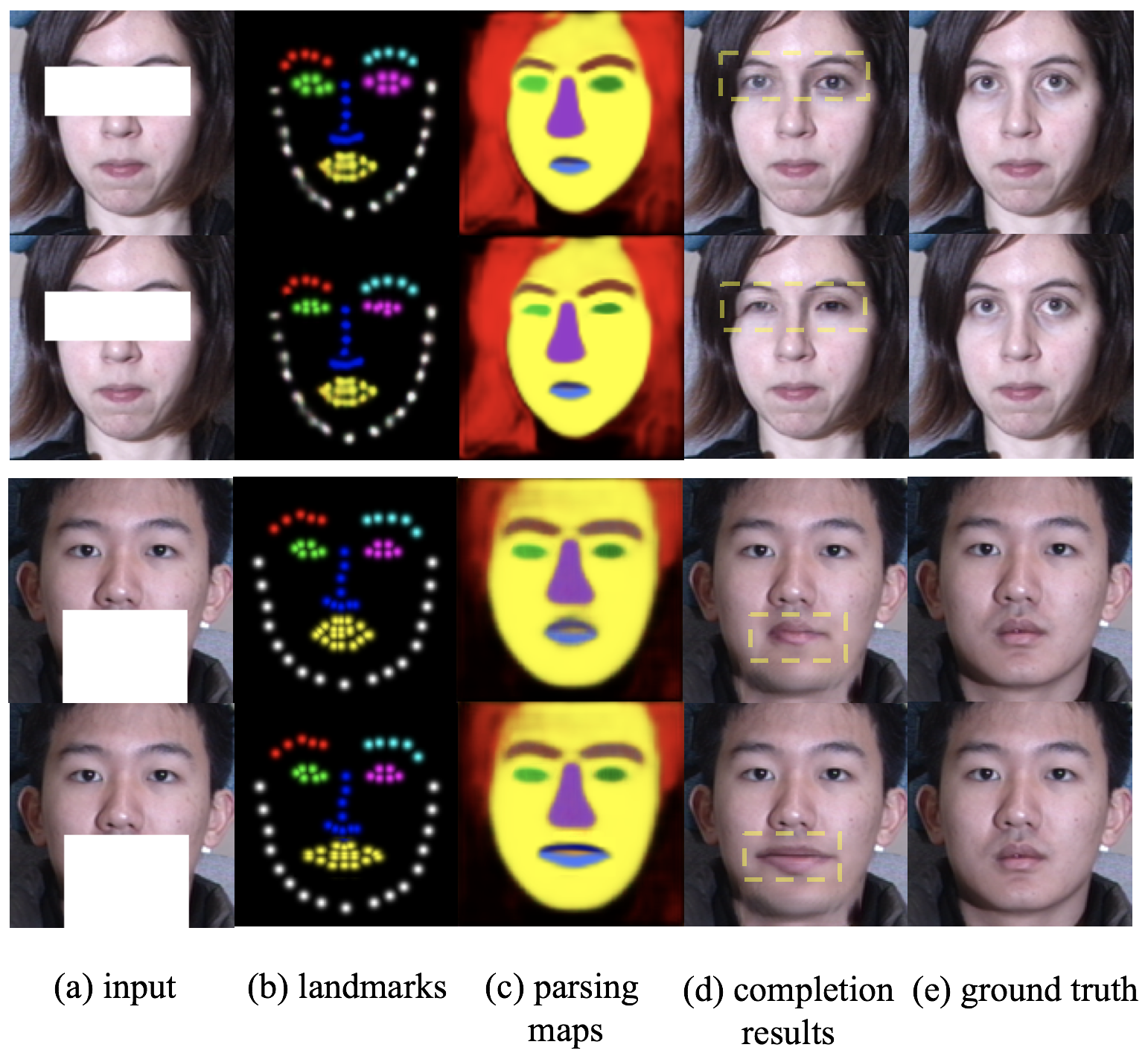}
  \caption{Face completion and attributes editing results. The first and third row show the visual quality results of face completion under the guidance of inferred facial geometry. In the second row, the girl's eyes become smaller after changing its facial geometry. In the fourth row, the boy's mouth becomes much bigger by replacing the mouth patch in the facial geometry images with a bigger mouth patch.}
  \label{demo}
\end{figure}

\section{Introduction}
Face completion, also known as face inpainting, aims to complete a face image with a masked region or missing content. As a common face image editing technique, it can also be used to edit face attributes. The generated face image can either be as accurate as the original face image, or content coherent to the context so that the completed image looks visually realistic. Most traditional methods \cite{barnes2009patchmatch,huang2014image,darabi2012image} rely on low-level cues to search for patches to synthesize missing content. These image completion algorithms are skilled in filling backgrounds which contain similar textures but not excel at specific image object like face since prior domain knowledge isn't well incorporated.

Recently, CNN based image completion methods with adversarial training strategy have already significantly boosted the image completion performance \cite{pathak2016context,yeh2017semantic,li2017generative}. These methods set a masked image as the network input to learn context features, then restore the missing content from the learned features. Some challenges still exist in these methods when they are applied to real-world face completion problems.  First of all, different from general objects, human faces have distinct geometry distribution, and hence face geometry prior is more likely to facilitate face completion \cite{yeh2017semantic}. Few existing methods utilize the facial prior knowledge and well incorporate it into an end-to-end network. Second, these algorithms are incapable of modifying the face attributes of the filled region \cite{li2017generative}, e.g., editing the eye shape or mouth size in figure \ref{demo}.
 
To address these two problems, this paper studies a geometry-aware face completion and editing network (FCENet) by exploring facial geometry priors including facial landmark heatmaps and parsing maps. Landmark heatmap is an image composed of 68 points labeled by different values to distinguish different face components. Similarly, facial parsing map is an image with different values representing different face components (e.g., eyes, nose, mouth, hair, cheek, and background). They can not only provide a hint of target face’s geometry information for face completion but also point out a novel way to modify face attributes of masked regions. The FCENet consists of three stages. In the first stage, facial parsing maps and landmark heatmaps are inferred from masked face images. In the second stage, we concatenate the masked image, inferred landmark heatmaps and parsing maps together as the input of a face completion generator to restore the uncorrupted face images and mask. In the third stage, two discriminators distinguish generated face images and real ones globally and locally to force the generated face images as realistic as possible. Furthermore, the low-rank regularization boosts the completion network to disentangle more complex mask from the corrupted face image. Our FCENet is efficient to generate face images with a variety of attributes depending on the facial geometry images concatenated with masked face image. A few face completion and editing examples of FCENet are shown in figure \ref{demo}.

The main contributions of this paper are summarized as follows:
\begin{enumerate}
\item We design a novel network called facial geometry estimator to estimate reasonable facial geometry from masked face images. Such facial geometry is leveraged to guide the face completion task. Several experiments systematically demonstrate the performance improvements from different facial geometry, including facial landmarks, facial parsing maps and both together.
\item The FCENet allows interactive facial attributes editing of the generated face image by simply modifying its inferred facial geometry images. Therefore, face completion and face attributes editing are integrated into one unified framework. 
\item Our face completion generator simultaneously accomplishes face completion and mask disentanglement from the masked face image. A novel low-rank loss regularizes the disentangled mask to further enhance similarity between the disentangled mask and the original mask, which enables our method to handle various masks with different shapes and sizes.
\item Experiments on the CelebA \cite{liu2015deep} and Multi-PIE \cite{gross2010multi} dataset demonstrate the superiority of our approach over the existing approaches.
\end{enumerate}

\begin{figure*}[htb]
  \centering
  \includegraphics[width=1.43\columnwidth]{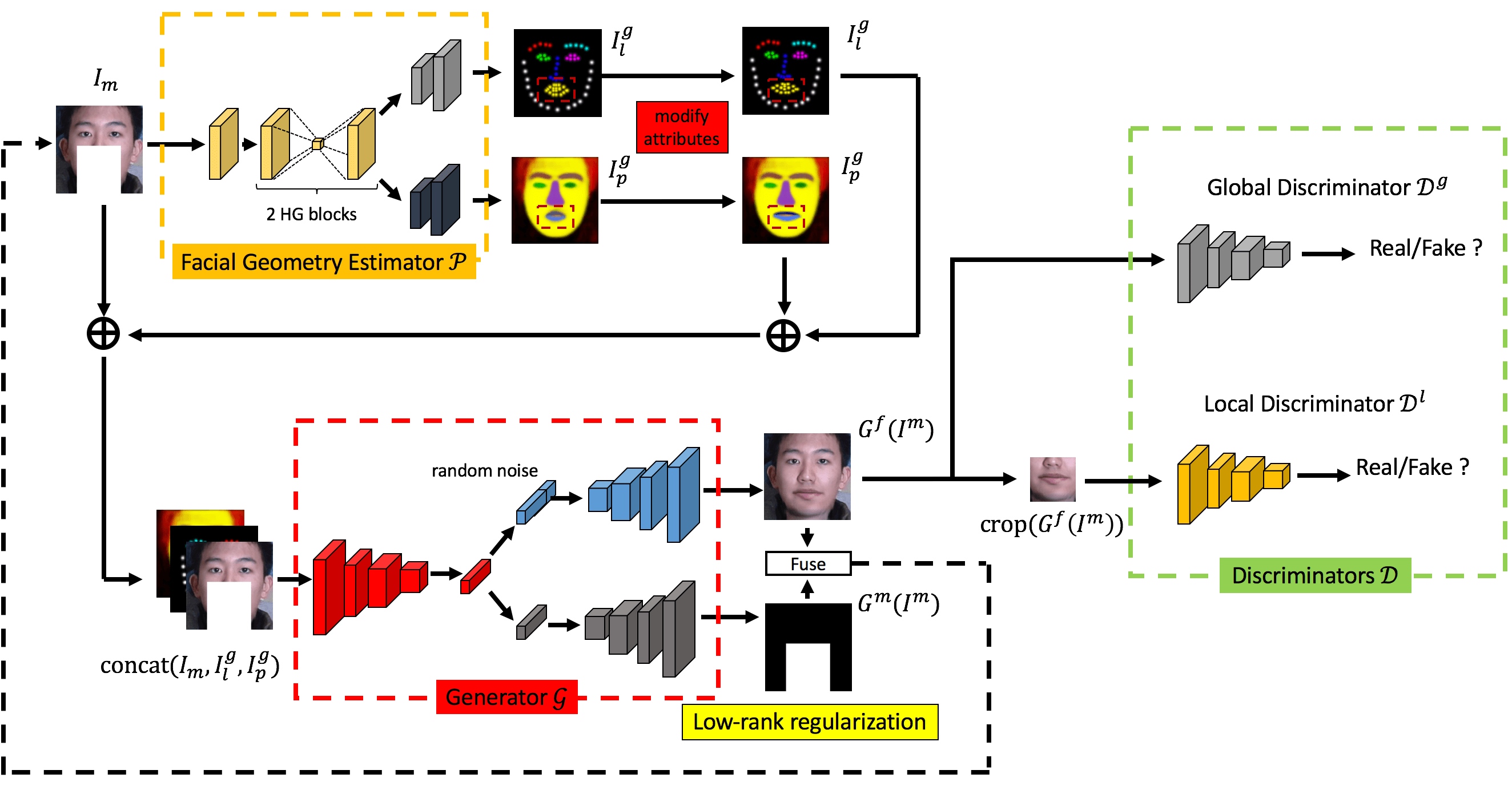}
  \caption{The overall architecture of the FCENet. It consists of three parts: a facial geometry estimator, a generator, and two discriminators. The facial geometry estimator infers reasonable facial geometry. The generator restores complete face images and disentangles masks, and the two discriminators distinguish real and generated complete face images globally and locally. The figure demonstrates that the boy's mouth is modified to be bigger by interactively changing his mouth in facial geometry images in the assumption that a certain user tends to edit the mouth attribute.}
  \label{overall_arch}
\end{figure*}

\section{Related Work}
\textbf{Generative Adversarial Network (GAN).} As one of the most significant image generation techniques, GAN \cite{goodfellow2014generative} utilizes the mini-max adversarial game to train a generator and a discriminator alternatively. The adversarial training method encourages the generator to generate realistic outputs to fool the discriminator. At the same time, the discriminator tries to distinguish real and generated images. In this way, the generator can generate samples that obey the target distribution and look plausibly realistic. Radford et al. propose a deep convolutional GAN (DCGAN) \cite{radford2015unsupervised} to generate high-quality images on multiple datasets. It is the first time to integrate the deep models into GAN. To address training instability of GAN, Arjovsky et al. \cite{arjovsky2017wasserstein} analyze the causes of training instability theoretically and propose Wasserstein GAN. Mao et al. \cite{mao2017least} put forward LSGAN to avoid vanishing gradients problem during the learning process. Recently, improved GAN architectures have achieved a great success on super-resolution \cite{ledig2016photo}, style transfer \cite{li2016combining}, image inpainting \cite{pathak2016context}, face attribute manipulation \cite{shen2017learning}, and face rotation \cite{huang2017beyond,Yibo2018pose}. Motivated by these successful solutions, we develop the FCENet based on GAN.

\noindent\textbf{Image Completion.} The image completion techniques can be broadly divided into three categories, The first category exploits the diffusion equation to propagate the low-level feature from the context region to the missing area along the boundaries iteratively \cite{bertalmio2000image,elad2005simultaneous}. These methods excel at inpainting small holes and superimposed text or lines but having limitations to the reproduction of large textured regions. The second category is patch-based methods, which observe the context of the missing content and search similar patch from the same image or external image databases \cite{darabi2012image,bertalmio2003simultaneous,criminisi2004region,hays2007scene}. These methods can achieve ideal completion results on backgrounds like sky and grass, but they fail to generate semantic new pixels if these patches don't exist in the databases. The third category is CNN-based, which train an encoder-decoder architecture network to extract image features from the context and generate missing content according to the image features \cite{pathak2016context,iizuka2017globally}. Deepak et al. \cite{pathak2016context} present an unsupervised visual feature learning algorithm called context encoder to produce a plausible hypothesis for the missing part conditioned on its surroundings. A pixel-wise reconstruction loss and an adversarial loss regularize the filling content to bear some resemblance to its surrounding area both on appearance and semantic segmentation. Satoshi et al. \cite{iizuka2017globally} put forward an approach which can generate inpainting images that are both locally and globally consistent by training a global and local context discriminator. Their method is able to complete images of arbitrary resolutions and various obstructed shapes by training a fully-convolutional neural network.

\noindent\textbf{Face Completion.} Human face completion is much more challenging than general image completion tasks because facial components (e.g., eyes, nose, and mouth) are of highly structurization and contain large appearance variations. Besides, the symmetrical structure is reflected on human faces like many natural objects. Second, compared to general object image completion, face completion need to pay more attention to preserving face identity. S. Zhang et al. \cite{zhang2018demeshnet,zhang2016multi} develop models to complete face images with structural obstructions like wavy lines. Y. Li et al. \cite{li2017generative} propose a generative face completion model assisted by a semantic regularization term. Their algorithm is sensitive to pose and expression variations and cannot manipulate face attributes of the missing areas. P. Liu et al. \cite{liu2017semantically} integrate perceptual loss into their network and replace the unmasked region of the generated face image with that from the original face image. Their approach produces high-quality face images with fine-grained details. Nevertheless, how to take advantage of facial geometry and control the face attributes of the filled region is still an open challenge, and that is the motivation of our FCENet.

\section{Proposed Method}
In this section, we introduce the FCENet for face completion and attributes editing. The FCENet consists of three parts: first, a facial geometry estimator learns to infer reasonable and natural facial parsing maps and landmark heatmaps. Second, an encoder-decoder structural generator restores the completion face image and disentangles the mask from the concatenation of the mask face image and facial geometry images. Third, global and local discriminators are introduced to determine whether the generated face images are sampled from ground truth distribution. The overall framework of our algorithm is shown in figure \ref{overall_arch}.

\begin{figure*}[htb]
  \centering
  \includegraphics[width=1.43\columnwidth]{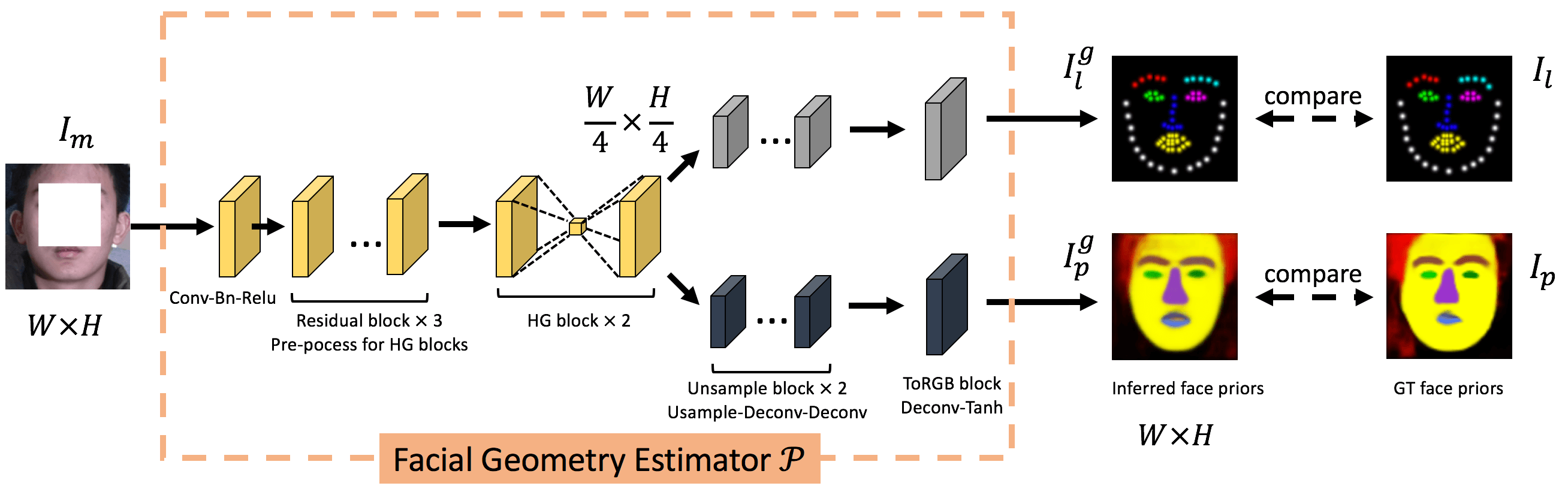}
  \caption{Detailed architecture of proposed facial prior estimator. The stacked hourglass blocks extract facial geometry features, the two branches infer natural facial landmark heatmaps and parsing maps based on shared facial geometry features.}
  \label{facial_geo_est}
\end{figure*}

\subsection{Network Architecture}\label{network_arch}
\subsubsection{Facial Geometry Estimator}
Geometry is the most distinct feature of the most real-world object, including human faces. Human faces contain various visual appearances due to factors like gender, illumination conditions, makeup, etc. However, similar facial geometry information still exists in these faces. As prior domain knowledge, facial geometry is rarely exposed to these influences. Most existing face completion algorithms have not yet explored the benefit of facial geometry prior. Thus we propose facial geometry estimator $\mathcal{P}$ to exploit facial geometry in face completion. The facial geometry serves as the guidance information for completion task, which is different from \cite{li2017generative} whose method treats the parsing map as a semantic regularization by closing the distance between the parsing maps of the original image and the generated image.

Details of the facial geometry estimator can be seen in figure \ref{facial_geo_est}. We apply the hourglass(HG) structure to estimate facial geometry inspired by its successful application on human pose estimation \cite{chen2017adversarial,newell2016stacked}. The HG block adopts a skip connection mechanism between symmetrical layers, which consolidates features across scales and preserves spatial information in different scales. 

\noindent\subsubsection{Face Completion Generator and Discriminators}
Our generator $\mathcal{G}$ has an encoder-decoder architecture that contains four parts: one encoder, two decoders, and one non-parametric fusion operation. Given the masked face images concatenated with inferred facial geometry images, the encoder extracts a feature vector that can be split into context feature vector and mask feature vector. These two feature vectors are fed into the face image decoder and mask decoder respectively. Our encoder and decoders have the same architecture, composing of 9 residual blocks evolved from ResNet \cite{he2016deep}. The structure of encoder is symmetrical to that of the two decoders, and they bear almost the same network architecture except for the input layer. By splitting the latent code inferred by the encoder into two feature vectors, face context feature and mask feature can be well disentangled. The generated face completion result is denoted as $G^f(I_m)$ and the disentangled mask is denoted as $G^m(I_m)$ where we omit $I_l^g$ and $I_p^g$ for simplicity. At last, the input masked face image is recovered as $I_r$ by a simple arithmetic operation conducted on $G^f(I_m)$ and $G^m(I_m)$ to facilitate the disentangling task further. In our framework, each pixel value is normalized between -1 and 1, and the disentangled mask is expected to be a matrix filled with -1 and 1, where 1 represents the masked area and -1 represents the unmasked area. Thus, the recovered masked face image is formulated as follows,
\begin{equation}
  I_r = \max(G^f(I_m),G^m(I_m))
\end{equation}
where the $\max(\cdot, \cdot)$ is an element-wise maximum function.

The generator completes face images under the guidance of inferred facial geometry from the masked input image. Our facial geometry-aware FCENet has already produced visually realistic face images in this way. However, some attributes of generated face images might not be satisfactory. Editing facial attributes by changing its guidance information is a natural idea, e.g., editing facial attributes by modifying inferred facial landmark heatmaps and parsing maps. Multiple ways can be explored in modifying facial geometry images. One alternative way is the copy-and-paste strategy, since abundant facial attributes exist in facial geometry images from the training set, thus image patches that possess desired facial attributes can be utilized. Another alternative way is modifying facial geometry images directly by moving landmark points or changing edges of parsing maps. We still denote modified facial landmark heatmaps and facial parsing maps as $I_l^g$ and $I_p^g$ respectively. Then under the guidance of modified facial geometry, our FCENet generates face images with desired facial attributes. Thus, by editing the inferred facial geometry of filled regions, as highlighted by the red box in figure \ref{overall_arch}. The generator is able to produce a diversity of face images with different attributes. Facial attributes editing cases are presented in section "Experiments and Analysis".

The generator can only capture coarse geometry and texture information of the filled region by optimizing the pixel-wise reconstruction loss between the original clean face $I$ and generated complete face $G^f(I_m)$. To capture fine details, the global and local discriminators are trained to compete with the generator.  By optimizing the generator and discriminators alternatively, the generator $\mathcal{G}$ produces complete face images which are photo-realistic and of high-quality\cite{li2017generative}. The global and local discriminators are composed of same network structure apart from the input layer. The confidence values they assign to the entire image and masked region patch are denoted as $D^g(\cdot)$ and $D^l(\cdot)$ respectively. The generator $\mathcal{G}$ is trained to generate visually realistic face image through adversarial training with the two discriminators, whose object is to solve the following min-max problem,
\begin{equation}
  \begin{aligned}
  \min_{\theta_G}\max_{\theta_{D^g}, \theta_{D^l}} &\mathbb{E}_{I\sim P(I)}[ \log D^g(I) + \log D^l({\rm crop}(I)) ] +\\
  & \mathbb{E}_{I^m\sim P(I^m)}[ \log(1-D^g(G^f(I^m))) + \\
  & \log(1-D^l({\rm crop}(G^f(I^m))))]
  \end{aligned}
\end{equation}
where we denote ${\rm crop}(\cdot)$ as the function to crop the patch of the original missing region from the generated complete face image.

\subsection{Loss Functions}\label{losses}

\begin{figure}[htb]
  \centering
  \includegraphics[width=.86\columnwidth]{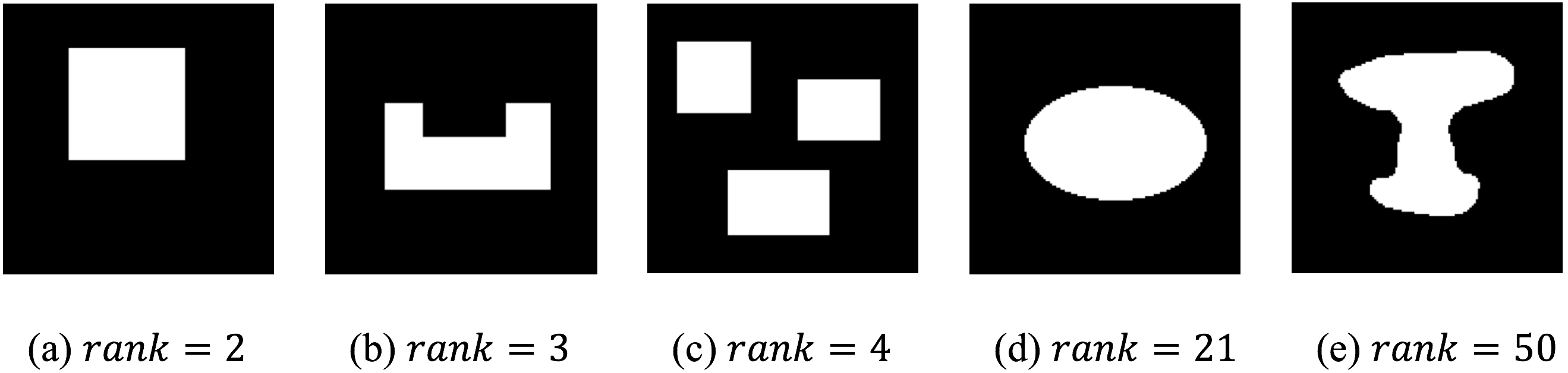}
  \caption{Ranks of some typical mask matrices. The size of the above matrices is $128\times 128$. In these matrices, black represents -1 and white represents 1. From (a) to (e) we notice that the more complex the mask is, the larger its rank value will be. Thus low-rank loss regularizes disentangled masks since the human labeled mask won't be much too complicated.}
  \label{rank}
\end{figure}

\textbf{Low-rank Regularization.}
The mask image we adopt in our approach is a black-background gray image with white squares representing masked regions. Thus, the mask matrix contains only 1 and -1, where 1 for the masked region and -1 for the unmasked region. The rank values of several typical mask matrices are presented in figure \ref{rank}. Therefore, low-rank regularization is beneficial to the denoising of the disentangled mask matrix. When the low-rank regularization is incorporated into the proposed FCENet, it is required to be back-propagated, whereas simple $rank(\cdot)$ function doesn’t satisfy such a condition. Thus, an approximate variant is applied in our algorithm in practice. We notice that if the mask matrix $M$ is a low-rank matrix, its elements are linearly correlated. Then $M^{\rm T}M$ tends to be a block-diagonal matrix. Thus, its nuclear norm can be used to measure its rank, i.e.

\begin{equation}
 ||M||_*={\rm tr}(\sqrt{M^{\rm T}M})
\end{equation}

\noindent\textbf{Pixel-wise Reconstruction Loss.}
In our approach, the pixel-wise loss is adopted to accelerate optimization and boost the superior performance. For the generated image $I^{gen}$ and the target image $I^{target}$, the pixel-wise loss is written as:
\begin{equation}
  L_{rec}=\frac{1}{W\times H}\sum_{x=1}^{W}\sum_{y=1}^{H}|I_{x,y}^{gen}-I_{x,y}^{target}|^p
\end{equation}
where $p$ is set as 1 for $L1$ loss or 2 for $L2$ loss. The pixel-wise reconstruction loss is measured at five generated image: the inferred facial geometry images $I_l^g$ and $I_p^g$, the generated complete face $G^f(I_m)$, the disentangled mask $G^m(I_m)$, and the reconstructed masked face image $I_r$.

\noindent\textbf{Adversarial Loss.}
The adversarial losses for both the complete face image and its masked region are calculated. The adversarial loss pushes the synthesized images to reside in the manifold of target image distribution. We denote $N$ as batch size, the adversarial loss for the complete face image is calculated as:
\begin{equation}
  L_{adv}^{g} = \frac{1}{N}\sum_{n=1}^{N}-\log D^g(G^f(I_m))
\end{equation}
Similarly, the adversarial loss for missing region is formulated as:
\begin{equation}
  L_{adv}^{l} = \frac{1}{N}\sum_{n=1}^{N}-\log D^l({\rm crop}(G^f(I_m)))
\end{equation}
The overall adversarial loss is:
\begin{equation}
  L_{adv} = \alpha L_{adv}^{g} + \beta L_{adv}^{l}
\end{equation}
The parameters $\alpha$ and $\beta$ are set as 1.

\noindent\textbf{Symmetric Loss.}
Symmetry is the prior geometry knowledge widely existing in human faces, especially reflected on frontal face images. To preserve such a symmetrical structure, a symmetric loss is imposed to constraint the synthesized face images and accelerate the convergence of our algorithm. For a face completion result $I^g$, its symmetric loss takes the form:
\begin{equation}
  L_{sym}=\frac{1}{W/2\times H}\sum_{x=1}^{W/2}\sum_{y=1}^{H}|I_{x,y}^g-I_{W-(x-1),y}^g|
\end{equation}
However, the limitation of the above pixel-level symmetric loss is obvious and can be divided into three folds: the illumination changes, the intrinsic texture difference, and human face poses. Hence its weight in the overall loss is not heavy.

\noindent\textbf{Overall Loss.}
The final objective loss to be optimized is the weighted sum of all the loss mentioned above:
\begin{scriptsize}
\begin{equation}
    L= \lambda_1 L_{rec}^{gen} + \lambda_2 L_{rec}^{geo} + \lambda_3 L_{adv} + \lambda_4 L_{rank} + \lambda_5 L_{sym} + \lambda_6 L_{tv}
\end{equation}
\end{scriptsize}
where $L_{tv}$ is a regularization term on generated face images to reduce spiky artifacts \cite{johnson2016perceptual}. $L_{rec}^{gen}$ and $L_{rec}^{geo}$ represent the pixel-wise reconstruction losses for generated image and geometry image respectively.

\section{Experiments and Analysis}
\subsection{Experimental Settings}
\textbf{Datasets.} We evaluate our model under both controlled and in-the-wild settings. To this end, two publicly available datasets are employed in our experiments: Multi-PIE \cite{gross2010multi} and CelebA \cite{liu2015deep}. The Multi-PIE is established for studying on the PIE (pose, illumination, and expression) invariant face recognition. It consists of 345 subjects captured in controlled environments. The Multi-PIE dataset contains face images with variational face poses, illumination conditions, and expressions. We choose face images with the frontal view and balanced illumination, resulting in 4539 images of 345 subjects. We use images from the first 250 subjects for training and the rest for testing, and there is no overlap between the training and testing sets. Thus, the training set contains 3627 images belonging to 250 individuals, and the testing set contains 912 images belonging to 95 individuals. The face regions are cropped by their eyes locations and resized to $128\times 128$. 

CelebA consists of 202,599 celebrity images with large variations in facial attributes. These images are obtained from unconstrained environments. The standard split for CelebA is employed in our experiments, where 162,770 images for training, 19,867 for validation and 19,962 for testing. Following \cite{radford2015unsupervised}, we crop and roughly align all the face images by the locations of the centers of eyes.

\noindent\textbf{Implementation Details.} We use colorful images of size $128\times 128\times 3$ in all the experiments. The width and height of rectangle masks are randomly selected with mean shape $64\times 64$ for training so that at least one sense organ is under the obstruction. Whereas, the position is randomly selected to prevent the model from putting too much attention on completing a certain facial part. The images are random flipped horizontally by probability 0.5. We set the learning rate as 0.0002 and deploy Adam optimizer \cite{kingma2014adam} for the facial geometry estimator, the generator and the two discriminators. The FCENet is trained for 200 epochs on the Multi-PIE and 20 epochs on the CelebA. The weights $\lambda_1,\ \lambda_2,\ \lambda_3,\ \lambda_4,\ \lambda_5,\ \lambda_6$ in overall loss are set as $10, 1, 1, 0.001, 0.01, 0.0001$ in practice, respectively. We apply end-to-end training in FCENet and facial attributes manipulation is only conducted on testing phase.

\noindent\textbf{Ground Truth Facial Geometry Extractors.} Inferring facial landmark heatmaps and parsing maps is a significant step in the FCENet. In our algorithm, Facial heatmaps of 68 landmarks and parsing maps of 11 components are used to supervise the facial geometry estimator. But these two facial geometry information is not provided in the Multi-PIE and CelebA datasets. Therefore, we deploy open source state-of-the-art face alignment \cite{bulat2017far} and face parsing \cite{Liu_2015_CVPR} tools to extract facial landmarks and parsing maps from original complete face images as the ground truth geometry, then the facial geometry estimator is trained to recover these ground truth facial geometry from masked input face images.

\subsection{Comparison Results}
To demonstrate the effectiveness of our algorithm, we make comparisons with several state-of-the-art methods: PatchMatch (PM) \cite{barnes2009patchmatch}, Context Encoder (CE) \cite{pathak2016context}, Generative Face Completion (GFC) \cite{li2017generative} and Semantic Image Inpainting (SII) \cite{yeh2017semantic}. The performances are evaluated both qualitatively and quantitatively. Specifically, three evaluation metrics are considered: visual quality, Peak Signal-to-Noise Ratio (PSNR) and Structural SIMilarity (SSIM). PSNR directly measures the difference in pixel values and SSIM estimates the holistic similarity between two images.

\begin{table}[htbp]
  \centering
  \caption{Quantitative results on the Multi-PIE and CelebA testing sets. Bold type indicates the best performance.}
    \begin{tabular}{ccccc}
    \toprule
    Dataset & \multicolumn{2}{c}{Multi-PIE} & \multicolumn{2}{c}{CelebA} \\
    \midrule
    Method & PSNR  & SSIM  & PSNR  & SSIM \\
    \midrule
    % PM    & 30.160     & 0.909     & 29.130     & 0.896 \\
    CE    & 23.052     & 0.678     & 24.499     & 0.732 \\
    SII   & 19.366     & 0.682     & 18.963     & 0.685 \\
    GFC   & 27.208     & 0.889     & 24.281     & 0.836 \\
    Ours  & \textbf{27.714} & \textbf{0.904} & \textbf{24.944} & \textbf{0.871} \\
    \bottomrule
    \end{tabular}%
  \label{PSNR/SSIM}%
\end{table}%

\noindent\textbf{Quantitative Results.} Comparisons on PSNR and SSIM are shown in table \ref{PSNR/SSIM}. For fair comparisons, we retrain the CE model on the Multi-PIE and CelebA since it is not trained for face completion, and the GFC model is retrained on the Multi-PIE. The publicly available SII implementation is trained only on CelebA and doesn't support training on other datasets. We find these indexes are basically consistent with the visual quality, i.e., the more plausible visual quality, the higher PSNR and SSIM. On quantitative results, our FCENet outperforms SII, CE, and GFC. Compared to general image completion models(CE, SII), performances of models considering facial domain knowledge(GFC, ours) are substantially better. We also investigate performance improvement from GFC to FCENet by comparing their networks. Our FCENet exploits facial geometry as guidance information for face completion while GFC treats them as semantic regularization. More valid facial geometry information is utilized by feeding them into the face completion generator than by regularizing face semantic loss. Our FCENet makes use of facial landmark heatmaps and parsing maps that carry richer facial geometry information than GFC which only deploys facial parsing maps.

\begin{figure}[htb]
  \centering
  \includegraphics[width=.71\columnwidth]{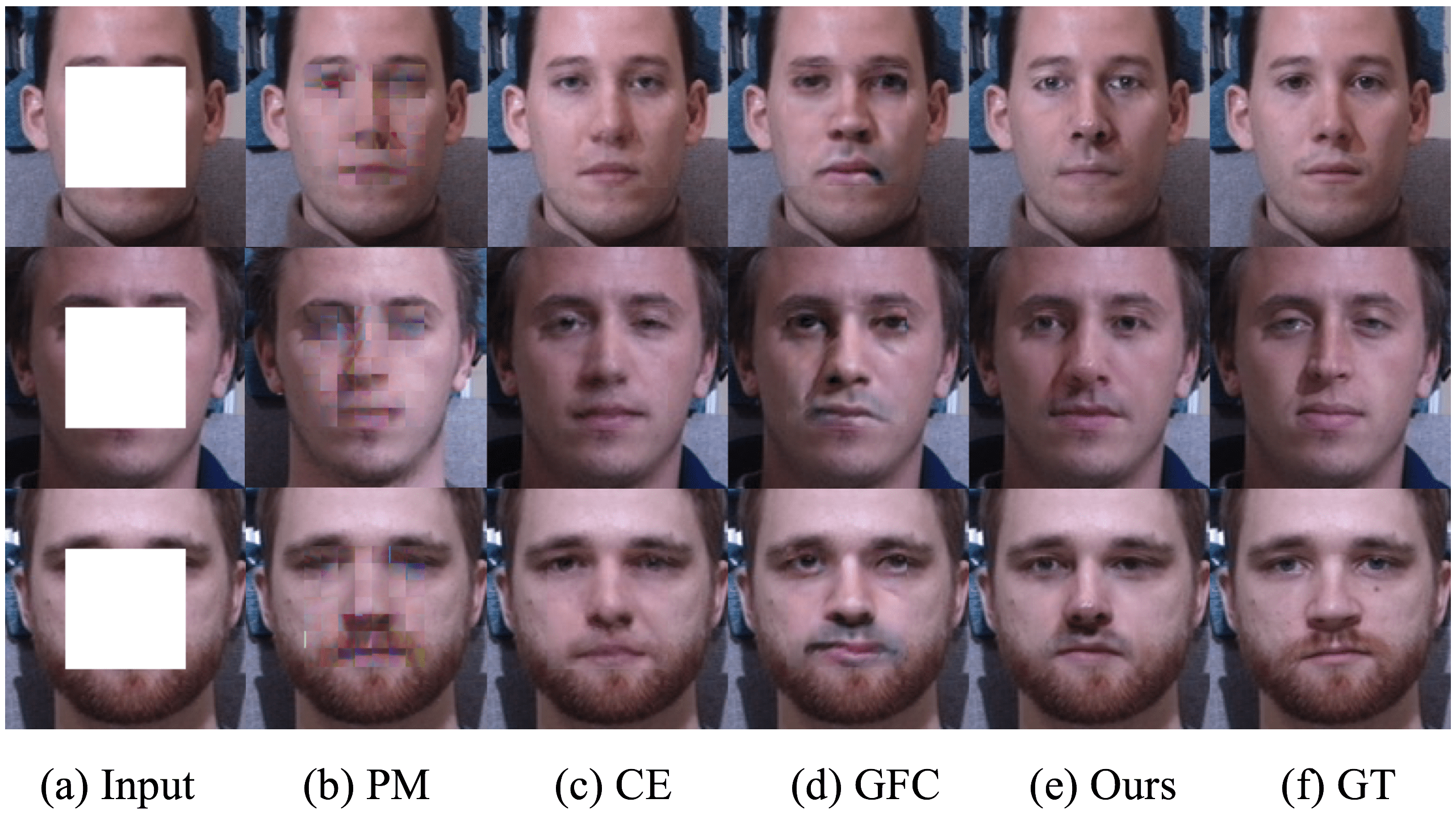}
  \caption{Visual quality of completion results on the Multi-PIE. Our completion results vastly outperform other methods on visual quality.}
  \label{comp_multipie}
\end{figure}

\begin{figure}[htb]
    \includegraphics[width=.86\columnwidth]{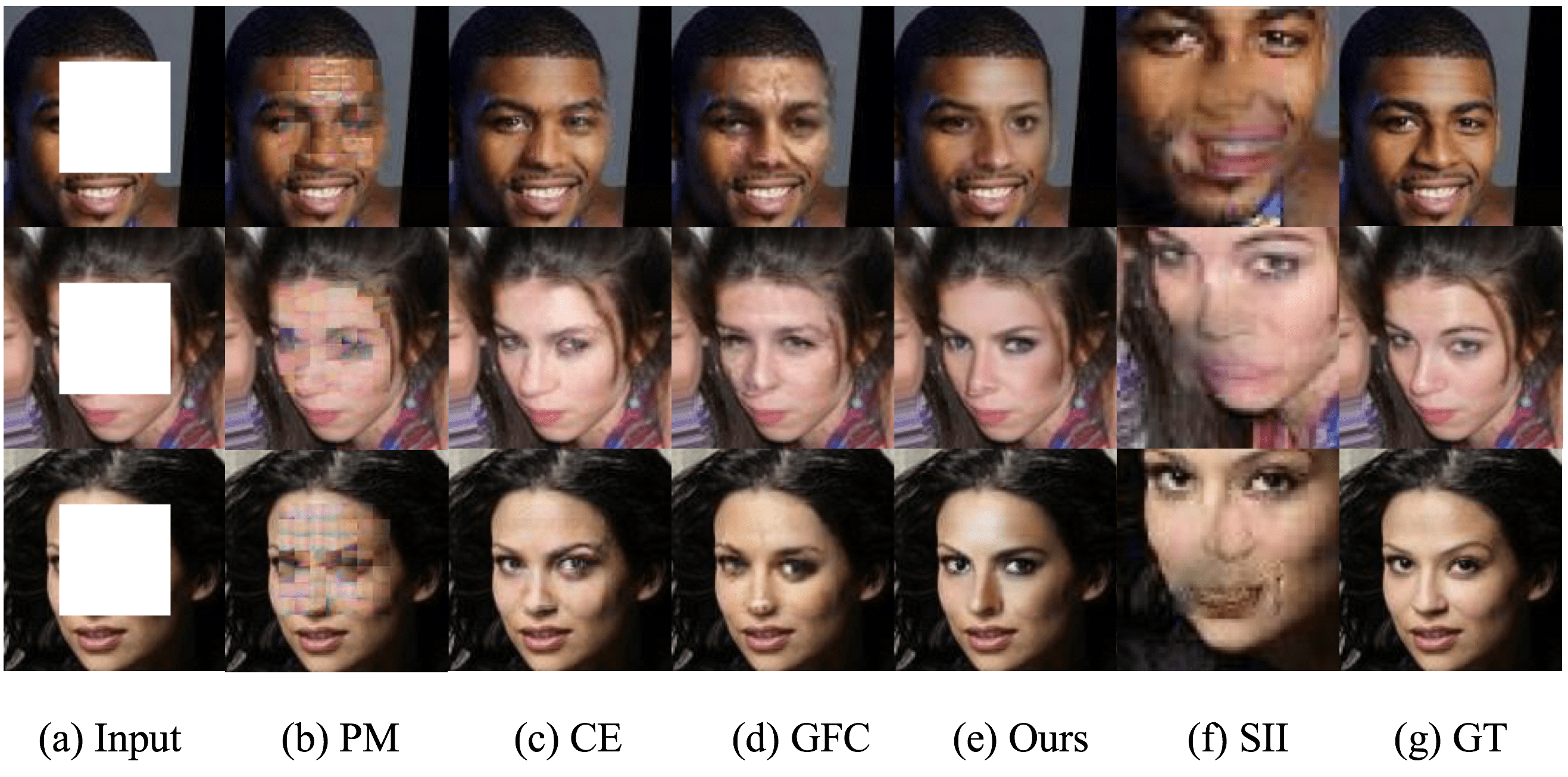}
    \caption{Visual quality of completion results on the CelebA. For SII, we use its public implementation on CelebA to complete face images. Note that SII completes face images cropped at the center to $64\times 64$.}
    \label{comp_celeba}
\end{figure}

\begin{figure}[htb]
  \centering
  \includegraphics[width=.71\columnwidth]{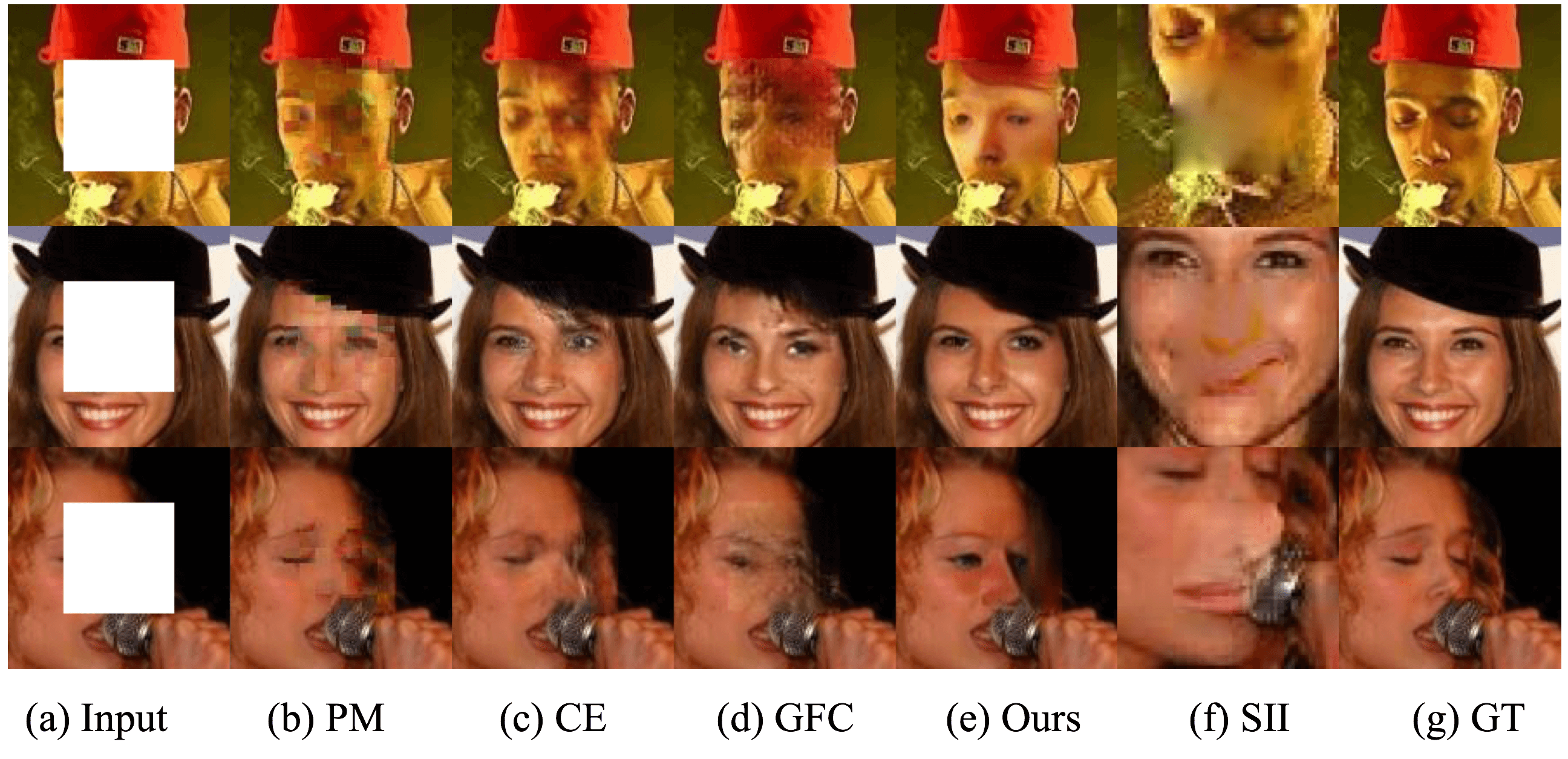}
  \caption{Hard cases selected from the CelebA. The face images in the three rows pose difficulty of exceptional context(red background), inpainting face with occlusion(hat) and posed face. Our FCENet generates visually realistic face images while reconstruction artifacts can be easily observed in PM, SII, CE and GFC.}
  \label{hard}
\end{figure}

\noindent\textbf{Qualitative Results.} The visual comparisons are summarized in figure \ref{comp_multipie} and \ref{comp_celeba}. We mainly show typical examples with masks covering key face components (e.g., eyes, mouths, and noses). The PM method searches the most similar patches from the masked input and fails to complete appealing visual results. It is not surprising because the missing parts differ from the rest image regions in pixel level and PM does not consider the semantic relevance. SII generates relatively reasonable results but the completion results lack global consistency with the original input. About CE and GFC, reconstruction artifacts can be observed especially for those hard cases as shown in figure \ref{hard}. Overall, our model can generate the most realistic results thanks to the guidance of the geometry information about facial anatomy.

\subsection{Ablation Study}

\noindent\subsubsection{Effects of Low-rank Regularization and Different Facial Geometry Images}

To validate the effects of the low-rank loss, estimated facial landmark heatmaps and facial parsing maps. Different experiment settings are applied as table \ref{exp_set_abl} presents.
\begin{table}[htbp]
  \centering
  \footnotesize
  \caption{Different experiment settings and their testing results (PSNR, SSIM) on the Multi-PIE for ablation study, higher values are better.}
     \resizebox{.95\columnwidth}{!}{
     \begin{tabular}{cccccc}
    \toprule
    Exp. Setting & 1 & 2 & 3 & 4 & 5 \\
    \midrule 
    Low-rank Loss & $\times$ & $\checkmark$ & \checkmark & \checkmark & \checkmark \\
    Landmarks & $\times$ & $\times$ & \checkmark & $\times$ & \checkmark \\
    Parsing Maps & $\times$ & $\times$ & $\times$ & \checkmark & \checkmark \\
    PSNR & 24.926 & 25.470 & 27.256 & 27.482 & \textbf{27.714} \\
    SSIM & 0.841 & 0.845 & 0.898 & 0.900 & \textbf{0.904} \\
    \bottomrule
    \end{tabular}}
  \label{exp_set_abl}%
\end{table}%

\begin{figure}[htb]
  \centering
  \includegraphics[width=.71\columnwidth]{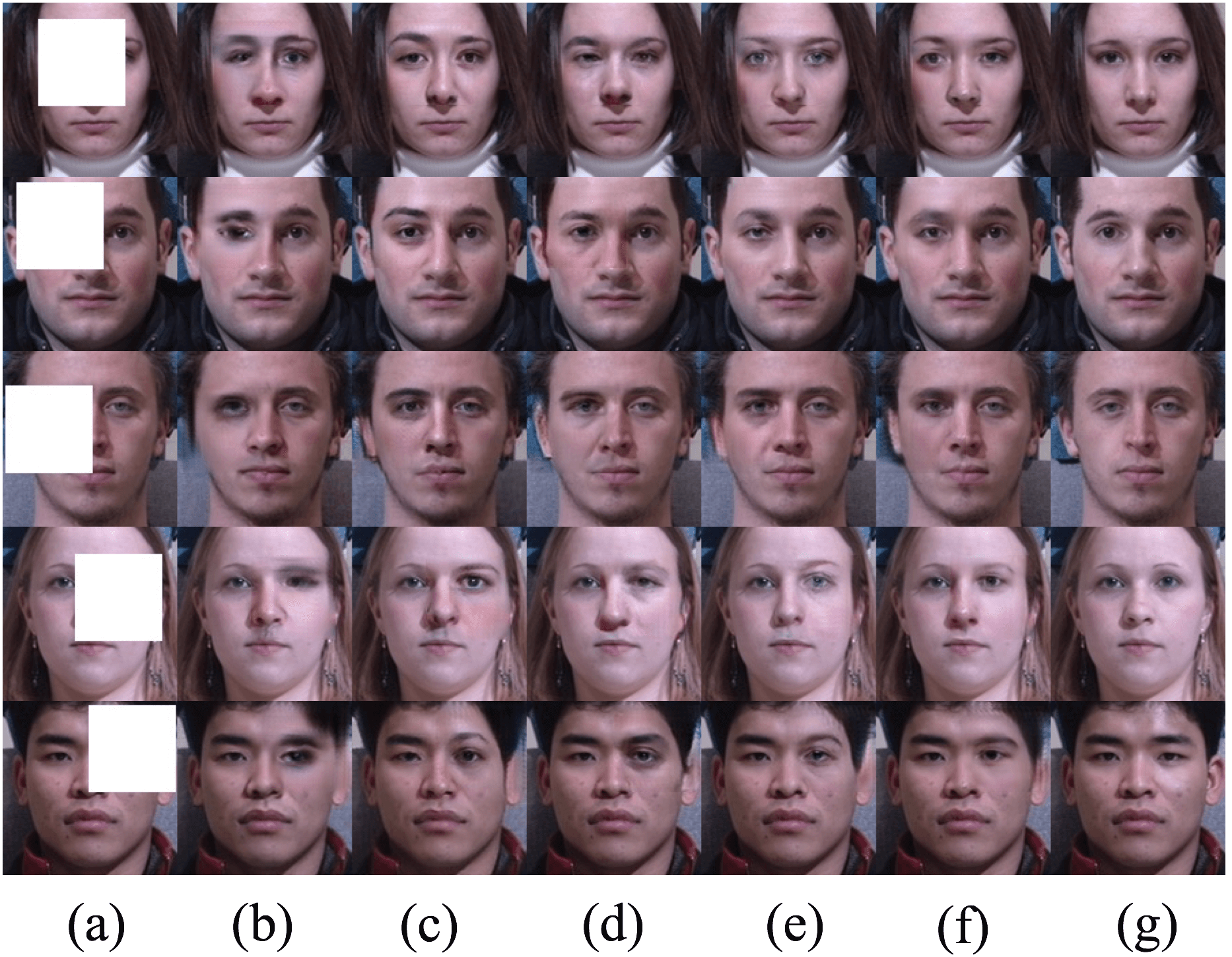}
  \caption{Visual quality of completion results under different experiment settings. (a) and (g) are masked input and ground truth complete face images respectively, (b)-(f) are completion results on the Multi-PIE testing set under experiment setting 1-5 respectively. Effects of low-rank regularization term and facial geometry are presented intuitively.}
  \label{self_compare}
\end{figure}

The experiment setting 1 and 2 are used to verify the effect of the low-rank regularization. The experiment setting 3, 4 and 5 respectively reflect the performance boosting of facial landmark heatmaps, facial parsing maps as well as both together in comparison to experiment setting 2 that doesn’t contain any facial geometry. Two metrics, PSNR and SSIM are calculated at the testing set as shown in table \ref{exp_set_abl}. The visual quality of testing results under different experiment settings is demonstrated as figure \ref{self_compare}.

\begin{figure}[htb]
  \centering
  \includegraphics[width=.71\columnwidth]{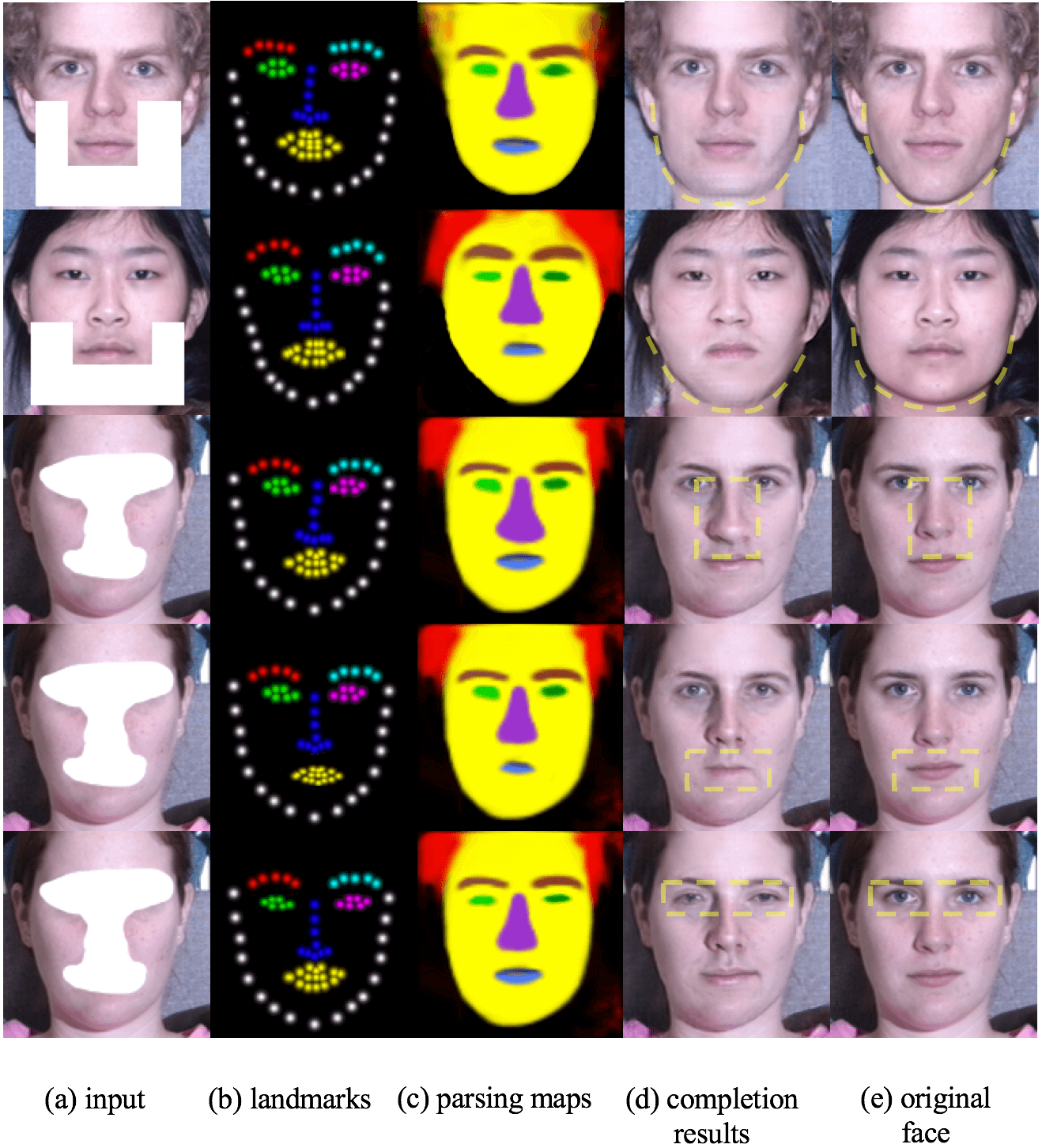}
  \caption{Attributes editing results. (b) and (c) are modified facial geometry. The face shapes in the 1st and 2nd rows are changed. Row 3, 4 and 5 present that nose, mouth and eyes are edited respectively while other attributes are nearly unchanged.}
  \label{attr_edit}
\end{figure}

\subsection{Face Attributes Manipulation}\label{attr_edit_txt}
After filling the masked regions, users may not be satisfied with the face attributes of these parts and tend to modify them. Our estimated facial geometry images allow the users to perform interactive face attributes editing on the resulting complete face. Examples are presented in figure \ref{demo}. Our algorithm also supports face attributes modifications for complete face images by manually pasting a white mask on the unsatisfied region of a complete face image. Then operations like copy-and-paste are used to modify facial geometry images with desired attributes. Hence we can simply change the facial geometry images to create novel human portraits. For example, changing the shape of generated faces by simply editing their geometry images (row 1 and 2 in figure \ref{attr_edit}), modifying one attribute while other attributes are kept similar to the target image (row 2, 3 and 4 in figure \ref{attr_edit}). 

\section{Conclusion}
This work has systematically studied facial geometry for face completion and editing, resulting in a new geometry-aware network, named FCENet. Our network is composed of a face geometry estimator for inferring facial geometry from a masked face, a generator for inpainting face images and disentangling their masks. A low-rank loss is imposed to regularize the mask matrix to denoise. Different from most existing face completion methods, our network naturally supports facial attributes editing by interactively modifying facial geometry images. Extensive experimental results on the widely used Multi-PIE and CelebA datasets demonstrate FCENet’s superior performance over state-of-the-art face completion methods.

\section{Acknowledgement}
This work is funded by National Natural Science Foundation of China (Grants No. 61622310), Youth Innovation Promotion Association CAS(2015190) and Innovation Training Programs for Undergraduates, CAS.

\bibliographystyle{aaai}
\bibliography{AAAI-SongL}

\end{document}